\documentclass{article}
\pdfoutput=1
\usepackage{spconf,amsmath}
\usepackage{graphicx}
\usepackage{amsfonts}

\usepackage{comment}
\usepackage{bm}
\usepackage{multirow}
\usepackage{here}
\usepackage{setspace}
\usepackage{color}
\title{Classifying degraded images over various levels of degradation\footnotemark}
\name{Kazuki Endo$^{\dagger}$ \qquad Masayuki Tanaka$^{\dagger \ddagger}$ \qquad Masatoshi Okutomi$^{\dagger}$}
\address{$^{\dagger}$ Tokyo Institute of Technology \\
    $^{\ddagger}$ National Institute of Advanced Industrial Science and Technology}

\graphicspath{{images/}}

\begin{document}
\maketitle

\begin{abstract}
Classification for degraded images having various levels of degradation is very important in practical applications.
This paper proposes a convolutional neural network to classify degraded images by using a restoration network and an ensemble learning.
The results demonstrate that the proposed network can classify degraded images over various levels of degradation well.
This paper also reveals how the image-quality of training data for a classification network affects the classification performance of degraded images.
\end{abstract}


\renewcommand{\thefootnote}{}
\footnotetext{\textcolor{blue}{\copyright 2020 IEEE. Personal use of this material is permitted. Permission from IEEE must be obtained for all other uses, in any current or future media, including reprinting/republishing this material for advertising or promotional purposes, creating new collective works, for resale or redistribution to servers or lists, or reuse of any copyrighted component of this work in other works. \\ This paper has been accepted by the 27th IEEE International Conference on Image Processing (ICIP 2020).}}
\renewcommand{\thefootnote}{\arabic{footnote}}
\setcounter{footnote}{0}

\section{Introduction}
Image degradation needs to be considered for a classification in practice.
Degraded images have each own level of degradation against clean images.
Thus, it is required for a classifier to classify degraded images having various levels of degradation.
Classification of degraded images has been investigated in several papers~\cite{L1,L3,L4,L5,L7,L8,L9,L10}.
Those papers have extended a classification of clean images based on a convolutional neural network(CNN)~\cite{D1,D2,D3,S5,D5,D6}.

\begin{figure}[t]
  \begin{minipage}[b]{1.0\linewidth}
    \centering
    \centerline{\includegraphics[clip,scale=0.50]{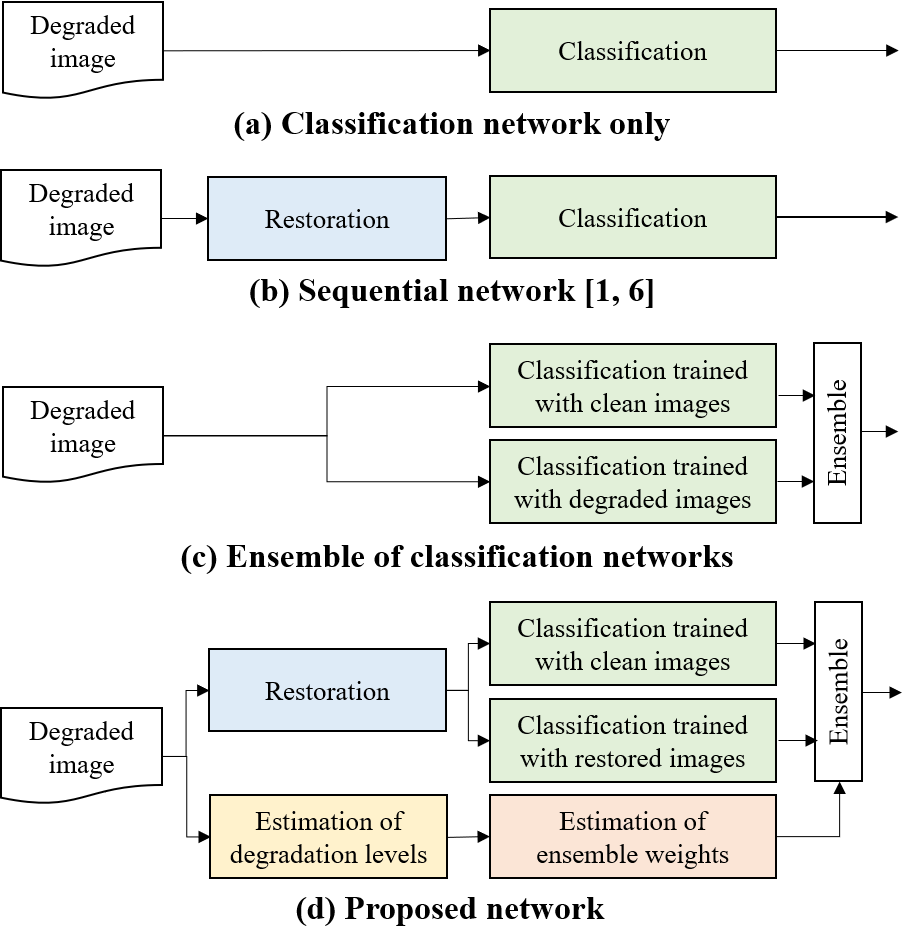}}
  \end{minipage}
  \vspace*{-1.00cm}
  \caption{Classification networks of degraded images}
  \label{ev_seq}
  \vspace*{-0.25cm}
\end{figure}

Figure~\ref{ev_seq}-(a) and~\ref{ev_seq}-(b) show two typical approaches to classify degraded images.
Figure~\ref{ev_seq}-(a) is a straightforward approach to train a classification network with degraded images, while the network architecture is identical to that for clean images.
Figure~\ref{ev_seq}-(b) is a sequential network which consists of an image restoration network and an image classification network.
These approaches can improve the classification performance of degraded images, especially for low-quality images.
However, they sometimes shows low performance for high-quality images~\cite{L3,L8}.
Figure~\ref{ev_seq}-(c) is an ensemble network of two classifiers trained with different image quality datasets; clean images and degraded images, which is described in~\cite{L7}.
In this paper, we propose a CNN-based classification network of degraded images which consists of an image restoration network and an ensemble network as shown in Fig.~\ref{ev_seq}-(d).
For the restoration, we adopt an existing high-performance CNN-based network~\cite{S1,S2,S3,S4,S6,S7,S8}.

Our main contributions are two points as follows.
The first point is to reveal how the classification performance of degraded images is affected by training data for a classification network, i.e. clean images or degraded images, and a restoration network used before a classification network.
The second point is that we propose an ensemble network of sequential networks, which can estimate suitable ensemble weights, as shown in Fig.~\ref{ev_seq}-(d).
Our proposed network outperforms the sequential networks for most of all degradation levels.

This paper focuses on JPEG distortion as an image degradation. 
However, the proposed method can be applicable to other image degradations.
We also applied it to the additive Gaussian noise. 

\section{Related works}
Classification of degraded images, such as low-resolution, noise, blurring, compression, etc., has been investigated ~\cite{L1,L3,L4,L5,L7,L8,L9,L10}.
Pei {\it et al.} have shown the impact of image degradation on the classification performance under several kinds of degradation~\cite{L8}.
Especially for haze and motion-blur, they have empirically shown that there are not much differences between the classification network only trained with degraded images and the sequential network incorporated a restoration~\cite{L4,L8}.
Endo {\it et al.} have proposed a classification network whose inputs are a JPEG image and a JPEG quality factor~\cite{L5}.
Gosh {\it et al.} have proposed an ensemble network of classification networks only trained with JPEG images~\cite{L7}.
They have also proposed a method based on maximum a posteriori (MAP) by using estimated JPEG quality factors and a simple method based on maximum probability.

Our proposed method has advantages induced by both a restoration network and an ensemble network for the classification of degraded images.

\begin{figure}[t]
  \begin{center}
    \includegraphics[clip,scale=0.40]{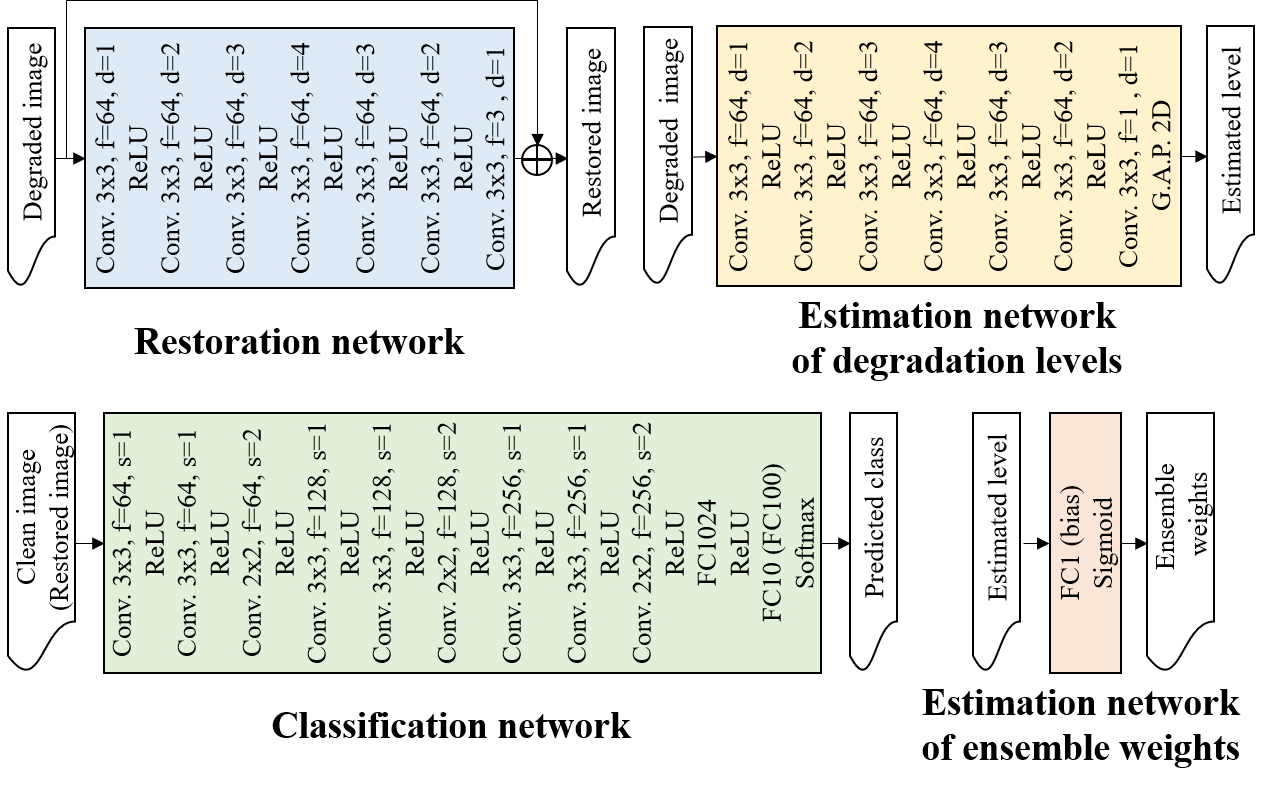}
  \end{center}
\vspace*{-0.75cm}  
\caption{Details of each network in the proposed network, where 3x3 or 2x2 represents the filter size, f is the dimension of feature map, d is the dilation rate, and s is the stride. ``G.A.P." denotes global average pooling. The classification network has two choices for training dataset; clean images and restored images.}
\label{architecture}
\vspace*{-0.25cm}
\end{figure}

\section{Proposed method}
\subsection{Proposed network}
Our proposed network is shown in Fig.~\ref{ev_seq}-(d).
The proposed network consists of four types of different networks; restoration network, classification networks, estimation network of degradation levels, and estimation network of ensemble weights.
Firstly, degraded images are restored by the restoration network.
The restored images are fed into two classification networks; the classification network trained with clean images and the classification network trained with restored images.
The classification networks infer each own probability vector.
On the other hand, degraded images are also fed into the estimation network of degradation levels.
Estimated degradation levels are fed into the estimation network of ensemble weights.
The estimation network of ensemble weights infers ensemble weights for two probability vectors predicted by two classification networks.
The weights take values in $[0,1]$, and the summation of the weights is one.
Finally, the predicted probability is calculated by weighted averaging.
 
Figure~\ref{architecture} shows the details of each network.
The restoration network is almost the same network as proposed in~\cite{S1}, where a batch normalization~\cite{D7} is omitted for the simplicity.
The estimation network of degradation levels is almost the same network as the network proposed in~\cite{L5,Q1}.
The classification network is a VGG-like network~\cite{D1}, where we use a spatial dropout~\cite{D3} and a convolution pooling~\cite{D2} instead of a max pooling. 
The optimizer is Adamax~\cite{D4} for all the training.

\subsection{Training procedure}
Four types of networks are trained separately.
Firstly, the restoration network is trained with pairs of degraded images and clean images, where its loss function is the mean square error (MSE) between clean images and restored images.
Degraded images are generated from clean images by applying some degradation operations.
Clean images are easily obtained from websites due to no need for any annotations. 

The estimation network of degradation levels is trained with pairs of degraded images and true degradation levels.
Its loss function is the MSE between true and predicted degradation levels.
Degraded images can be generated in the same way as in training the restoration network, where true degradation levels are known.

Two classification networks are trained with different data.
One is trained by using annotated clean images without any degradation.
Another one is trained by using restored images with annotation.
Degraded images with annotation are generated by applying some degradation to annotated clean images.
Restored images with annotation are acquired from the degraded images by using the restoration network, where all the weights of the restoration network are fixed during the training of the classification network.
Each loss function of the two networks is the cross-entropy between the correct labels and the predicted labels.

Finally, the estimation network of ensemble weights is trained by using degraded images with annotation, where its loss function is the cross-entropy between the correct labels and the predicted labels.
When the estimation network of ensemble weights is trained, all the weights of the other networks are fixed.

\section{Experiments}
Experiments are mainly focused on JPEG distortion.
We also show the results in the case of the additive Gaussian noise as another example of degradations.
The reproduction code is available on-line\footnote{http://www.ok.sc.e.titech.ac.jp/res/CNNIR/IRDI/}.

\subsection{Datasets and data augmentation}
Three datasets were used to train both the restoration network and the degradation level estimation network; Yang91~\cite{DataY}, Urban100~\cite{DataU}, and General100~\cite{DataG}.
We generated $64 \times 64$ sized patches from each image and applied data augmentation to them by using transpose, horizontal, and vertical flips.
Then, the JPEG compression was applied to the patches, where the JPEG quality factor was randomly sampled from 1 to 100\footnote{The details of the compression algorithm depend on the library. Python Image Library(PIL) was used for the JPEG compression. Note that the images compressed with the JPEG quality factor 100 also have the JPEG distortion.}.

The CIFAR datasets~\cite{DataC} were used to train the classification networks and the estimation network of ensemble weights.
Data augmentation was applied to the CIFAR images; zoom, shearing, horizontal flip, rotation, vertical, and horizontal shifts.
After that, the JPEG compression was also applied to each image in the same way mentioned above.
We denote these compressed CIFAR images as ``JPEG CIFAR".

\begin{table}[t]
  \small
  \caption{Comparison of networks in terms of incorporating a restoration and training data for a classification.}
  \label{comp_net}
  \vspace*{-0.25cm}
  \begin{center}
      \begin{tabular}{c|cccc} \hline
      \multirow{2}{*}{Name} & Cla & Cla & Seq & Seq \\
       & (org) & (jpg) & (org) & (res) \\ \hline
      Restoration & - & - & \checkmark & \checkmark \\ 
      Training data & Original & JPEG & Original & Restored \\ \hline
    \end{tabular}
  \end{center}
  \vspace*{-0.75cm} 
\end{table}

\begin{table}[t]
    \small
    \centering
    \caption{CPSNR[dB] of JPEG CIFAR-10 and its restoration. JPEG compression is applied to CIFAR-10 test images with each JPEG quality factor.}
    \vspace*{0.10cm} 
    \begin{tabular}{c|cccccc} \hline
         Quality factor & 10 & 30 & 50 & 70 & 90 \\ \hline
         JPEG & 23.28 & 26.73 & 28.25 & 29.82 & 33.65 \\
         Restoration & 24.29 & 27.91 & 29.45 & 31.00 & 34.51 \\  \hline 
    \end{tabular}
    \vspace*{-0.25cm} 
    \label{cpsnr}
\end{table}

\subsection{Interval mean accuracy}
We use an interval mean accuracy as a metric to evaluate the classification performance of images degraded with different degradation levels.
The following definition of the interval mean accuracy has been introduced in~\cite{L5}. 
\begin{equation*}
  \overline{Acc}\left({\bm \theta}; Q_l,Q_u\right)\stackrel{def}{=}\frac{\sum_{q=Q_l}^{Q_u}Acc\left({\bm f}({\rm D}\left({\bf X},q\right);{\bm \theta}), {\bf Y} \right)}{Q_u-Q_l+1},
\end{equation*}
where $\{Q_l,Q_u|Q_l < Q_u\}$ denote degradation levels, ${\rm D}\left({\bf X},q\right)$ is a degradation operator with a degradation level $q$ for clean images ${\bf X}$, 
${\bm f} (\cdot; {\bm \theta})$ represents classification network with parameters ${\bm \theta}$,
${\bf Y}$ represents true labels for ${\bf X}$, and $Acc$ is an accuracy.
The accuracy is a ratio dividing the number of predicted class labels, which coincide with correct class labels, by the number of all test samples.

\begin{table}[t]
  \small
  \caption{Interval mean accuracy of JPEG CIFAR-10.}
  \label{comp_net_cifar10}
  \begin{center}
      \begin{tabular}{c|cccccc} \hline
      \footnotesize{$\overline{Acc}$} of & Cla & Cla & Seq & Seq & Ens & Prop-\\
      $(Q_l,Q_u)$ & (org) & (jpg) & (org) & (res) & (Cla) & osed \\ \hline
      (1,20) & 0.431 & 0.724 & 0.569 & 0.736 & 0.677 & \bf{0.737} \\
      (21,40) & 0.700 & 0.844 & 0.802 & 0.852 & 0.825 & \bf{0.855} \\
      (41,60) & 0.763 & 0.857 & 0.849 & 0.864 & 0.844 & \bf{0.870} \\
      (61,80) & 0.799 & 0.866 & 0.874 & 0.870 & 0.858 & \bf{0.883} \\
      (81,100) & 0.861 & 0.874 & 0.902 & 0.878 & 0.885 & \bf{0.903} \\ \hline
      (1,100) & 0.711 & 0.833 & 0.799 & 0.840 & 0.818 &\bf{0.850} \\ \hline
    \end{tabular}
  \end{center}
  \vspace*{-0.55cm} 
\end{table}

\begin{figure}[t]
  \begin{center}
    \includegraphics[clip,width=8.5cm,height=5.00cm]{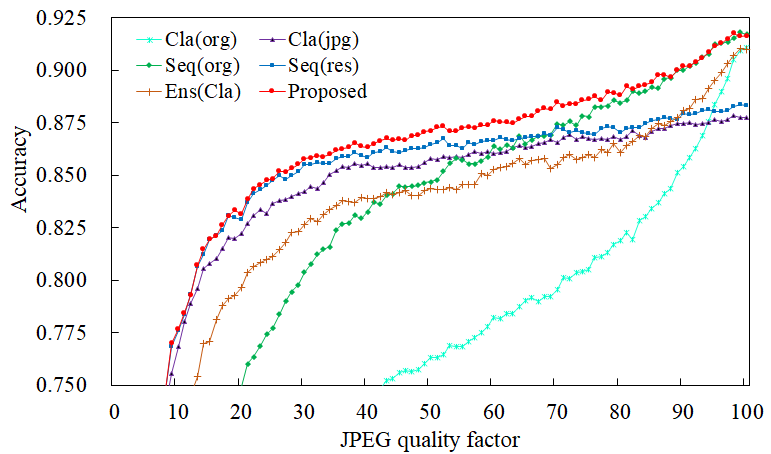}
  \end{center}
\vspace*{-0.75cm}
\caption{Accuracy of JPEG CIFAR-10.}
\label{gph_jpg_cifar10}
\end{figure}

\subsection{Performance analysis of classification networks for degraded images}
Here, we compare the performance of four networks summarized in Table~\ref{comp_net}.
``Cla(org)" and ``Cla(jpg)" are classification networks only trained by using original CIFAR-10 and JPEG CIFAR-10, respectively.
``Seq(org)" and ``Seq(res)" are sequential networks, where their classification networks are trained by using original CIFAR-10 and restored images of JPEG CIFAR-10, respectively.
Table~\ref{cpsnr} shows the color peak signal-to-noise ratio (CPSNR) of both JPEG CIFAR-10 images and images restored by the restoration network we used.

Figure~\ref{gph_jpg_cifar10} and Table~\ref{comp_net_cifar10} show the accuracy and the interval mean accuracy of JPEG CIFAR-10, respectively.
``Cla(org)" shows low performance under the existence of JPEG distortion.
``Seq(org)", which incorporates the restoration network before ``Cla(org)", outperforms ``Cla(org)" for all JPEG quality factors.
It shows that the restoration network helps ``Cla(org)" to classify degraded images.
However, ``Seq(org)" does not show enough performance for low quality factors when comparing to ``Cla(jpg)" which is directly trained with JPEG CIFAR-10.
``Cla(jpg)" shows roughly better performance than ``Cla(org)", but worse for the quality factors over around 95.
On the other hand, ``Seq(res)" slightly outperforms ``Cla(jpg)" for all JPEG quality factors, but still underperforms ``Cla(org)" for high quality factors.
When comparing ``Seq(res)" and ``Seq(org)", ``Seq(res)" is better than ``Seq(org)" for the quality factors under around 70, but worse over it.
Therefore, as for high-quality images, the classification network trained with clean images outperforms the classification networks trained with degraded images or restored images whether the restoration network is incorporated or not.

The best network is ``Seq(res)" for the quality factors under around 70, ``Seq(org)" for over it.
That is, the sequential networks outperform classification networks only. 

\begin{table}[t]
  \small
  \caption{Interval mean acurracy of JPEG CIFAR-100.}
  \label{comp_net_cifar100}
  \begin{center}
      \begin{tabular}{c|cccccc} \hline
      \footnotesize{$\overline{Acc}$} of & Cla & Cla & Seq & Seq & Ens & Prop- \\
      $(Q_l,Q_u)$ & (org) & (jpg) & (org) & (res) & (Cla) & osed \\ \hline
      (1,20) & 0.202 & 0.448 & 0.324 & 0.453 & 0.391 & \bf{0.454} \\
      (21,40) & 0.407 & 0.561 & 0.493 & 0.564 & 0.540 & \bf{0.572} \\
      (41,60) & 0.465 & 0.577 & 0.545 & 0.579 & 0.567 & \bf{0.594} \\
      (61,80) & 0.504 & 0.583 & 0.582 & 0.586 & 0.580 & \bf{0.611} \\
      (81,100) & 0.582 & 0.591 & 0.628 & 0.592 & 0.616 & \bf{0.637} \\ \hline
      (1,100) & 0.432 & 0.552 & 0.514 & 0.555 & 0.539 &\bf{0.573} \\ \hline
    \end{tabular}
  \end{center}
  \vspace*{-0.55cm}
\end{table}

\begin{figure}[t]
  \begin{center}
    \includegraphics[clip,width=8.5cm,height=5.00cm]{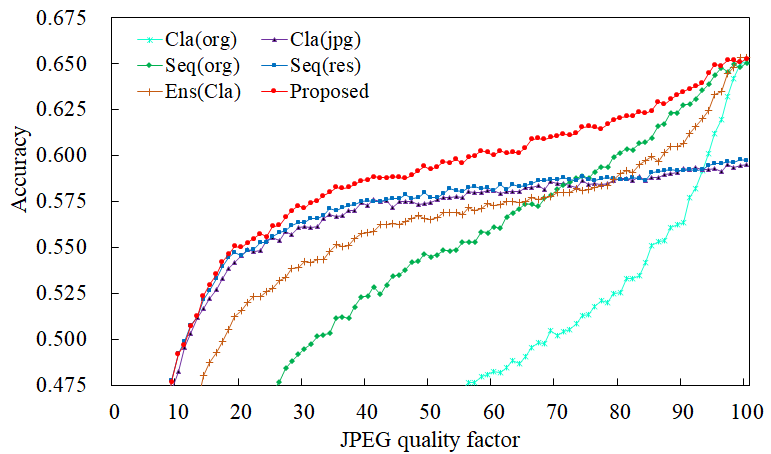}
  \end{center}
\vspace*{-0.75cm}  
\caption{Accuracy of JPEG CIFAR-100.}
\label{gph_jpg_cifar100}
\vspace*{-0.15cm}
\end{figure}

\subsection{Performance of the proposed network}
The proposed ensemble network, which is denoted as ``Proposed", uses two sequential networks; classification network trained with clean images and one trained with restored images.
These sequential networks could outperform classification networks only, as shown in the previous subsection.
Therefore, using these sequential networks is reasonable for ensemble learning.
An estimation network of JPEG quality factors was the same pre-trained network as reported in~\cite{L5}.
Figure~\ref{gph_jpg_cifar10} and Table~\ref{comp_net_cifar10} show that ``Proposed" outperforms both ``Seq(org)" and ``Seq(res)" in almost all JPEG quality factors.
That is, the proposed ensemble network can classify both high-quality and low-quality images well.
Moreover, ``Proposed" also outperforms ``Ens(Cla)" which denotes the ensemble network, as shown in Fig.~\ref{ev_seq}-(c), whose decision is taken by the maximum probability of ``Cla(org)" or ``Cla(jpg)".

Figure~\ref{gph_jpg_cifar100} and Table~\ref{comp_net_cifar100} show the accuracy and the interval mean accuracy of JPEG CIFAR-100, respectively.
``Proposed" also outperforms the other networks for the JPEG CIFAR-100.

\begin{table}[t]
   \small
    \centering
    \caption{CPSNR[dB] of Gaussian CIFAR-10 and its restoration. Gaussian noise is added to CIFAR-10 test images with each noise level.}
    \vspace*{0.10cm}
    \begin{tabular}{c|cccccc} \hline
         Noise level & 10 & 20 & 30 & 40 & 50 \\ \hline
         Gaussian & 28.13 & 22.11 & 18.59 & 16.09 & 14.15 \\
         Restoration & 33.47 & 29.74 & 27.58 & 26.05 & 24.87 \\  \hline 
    \end{tabular}
    \label{cpsnr_gauss}
    \vspace*{-0.25cm} 
\end{table}

\subsection{An example of other degradations}
\begin{figure}[!t]
  \begin{center}
    \includegraphics[clip,width=8.5cm,height=5.00cm]{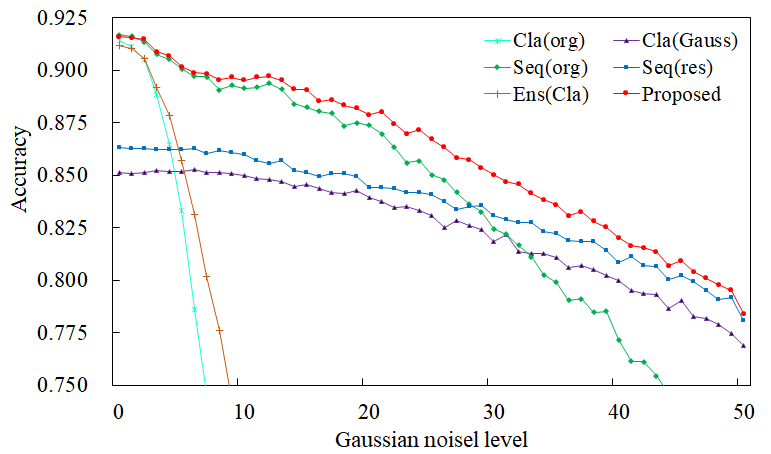}
  \end{center}
\vspace*{-0.75cm}  
\caption{Accuracy of Gaussian CIFAR-10.}
\label{gph_noise_cifar10}
\vspace*{-0.15cm}
\end{figure}

We applied the proposed method for the additive Gaussian noise.
Degradation operator was just replaced from JPEG compression to Gaussian noise.
We added CIFAR-10 images and Gaussian noise whose noise level changed from 0 to 50 for the 8-bit image.
We call the images ``Gaussian CIFAR-10".
Regarding the estimation network of the Gaussian noise level, the same pre-trained network, as reported in \cite{L5}, was used.
Table~\ref{cpsnr_gauss} shows the CPSNR of the restoration network.
Figure~\ref{gph_noise_cifar10} and Table~\ref{comp_net_nosie_cifar10} show the accuracy and the interval mean accuracy of Gaussian CIFAR-10, respectively.
``Proposed" almost outperforms the other networks for the Gaussian CIFAR-10.

\begin{table}[t]
  \small
  \caption{Interval mean accuracy of Gaussian CIFAR-10.}
  \vspace*{-0.25cm} 
  \label{comp_net_nosie_cifar10}
  \begin{center}
      \begin{tabular}{c|cccccc} \hline
      \footnotesize{$\overline{Acc}$} of & Cla & Cla & Seq & Seq & Ens & Prop- \\
      $(Q_l,Q_u)$ & (org) & (Gauss) & (org) & (res) & (Cla) & osed \\ \hline
      (0,10) & 0.796 & 0.851 & 0.903 & 0.862 & 0.838 & \bf{0.904} \\
      (11,20) & 0.328 & 0.844 & 0.883 & 0.852 & 0.509 & \bf{0.889} \\
      (21,30) & 0.138 & 0.829 & 0.848 & 0.838 & 0.271 & \bf{0.865} \\
      (31,40) & 0.106 & 0.809 & 0.797 & 0.821 & 0.198 & \bf{0.835} \\
      (41,50) & 0.101 & 0.785 & 0.740 & 0.798 & 0.173 & \bf{0.804} \\ \hline
      (0,50) & 0.304 & 0.824 & 0.835 & 0.835 & 0.406 & \bf{0.860} \\ \hline
    \end{tabular}
  \end{center}
  \vspace*{-0.55cm} 
\end{table}

\section{Conclusions}
\label{sec:conclusion}
We have proposed the ensemble network which shows higher performance for various degradation levels.
Firstly, we confirmed that two sequential networks, which are incorporating a restoration network into a classification network, outperform the classification networks only trained with clean or degraded images.
Then, we also found the sequential network shows the different performance depending on an image-quality of training data for classification networks.
Based on the results, the proposed network was constructed by using ensemble learning of the sequential networks, where the ensemble weights were inferred depending on the estimated degradation levels automatically.
Finally, we have shown that the proposed network is effective for not only the JPEG distortion but also the additive Gaussian noise.

\bibliographystyle{IEEEbib}
\begin{spacing}{0.85}
\bibliography{classify}
\end{spacing}
\end{document}